\documentclass[11pt,a4paper]{article}
\usepackage[hyperref]{acl2019}
\usepackage{times}
\usepackage{latexsym}
\usepackage{amsmath}
\usepackage{amssymb}
\usepackage{graphicx}
\usepackage{bm}
\usepackage{rotating,tabularx}

\usepackage{multirow}
\usepackage{url}

\usepackage{paralist}

\DeclareMathOperator\sigmoid{\operatorname{sigmoid}}

\aclfinalcopy 

\title{Hierarchical Transformers for Multi-Document Summarization}

\author{Yang Liu\and Mirella Lapata \\
    Institute for Language, Cognition and Computation
    \\
    School of Informatics, University of Edinburgh
    \\ 
    \texttt{yang.liu2@ed.ac.uk, mlap@inf.ed.ac.uk}
}

\date{}

\makeatletter
\newcommand{\thickhline}{%
    \noalign {\ifnum 0=`}\fi \hrule height 1pt
    \futurelet \reserved@a \@xhline
}
\makeatother

\begin{document}
    \maketitle
    \begin{abstract}
        
        In this paper, we develop a neural summarization model which can
        effectively process multiple input documents and distill
        abstractive summaries.  Our model augments a previously proposed
        Transformer architecture \cite{liu2018generating} with the
        ability to encode documents in a hierarchical manner.  We
        represent cross-document relationships via an attention
        mechanism which allows to share information as opposed to simply
        concatenating text spans and processing them as a flat sequence.
        Our model learns latent  dependencies among
        textual units, but can also  take advantage of explicit
        graph representations focusing on similarity or discourse
        relations.  Empirical results on the WikiSum dataset demonstrate
        that the proposed architecture brings substantial improvements
        over several strong baselines.\footnote{Our code and data is
            available at \url{https://github.com/nlpyang/hiersumm}.}

    \end{abstract}
    
    \section{Introduction}
    \label{sec:introduction}
    
    Automatic summarization has enjoyed renewed interest in recent
    years, thanks to the popularity of neural network models and their
    ability to learn continuous representations without recourse to
    preprocessing tools or linguistic annotations. The availability of
    large-scale datasets
    \cite{nytcorpus,hermann-nips15,newsroom-naacl18}
    containing hundreds of thousands of document-summary pairs has
    driven the development of neural architectures for summarizing
    \emph{single} documents. Several approaches have shown promising
    results with sequence-to-sequence models that encode a source
    document and then decode it into an abstractive summary
    \cite{see-acl17,asli-multiagent18,paulus2017deep,gehrmann2018bottom}.

    \emph{Multi}-document summarization --- the task of producing
    summaries from clusters of thematically related documents --- has
    received significantly less attention, partly due to the paucity
    of suitable data for the application of learning methods.
    High-quality multi-document summarization datasets (i.e.,~document
    clusters paired with multiple reference summaries written by
    humans) have been produced for the Document Understanding and Text
    Analysis Conferences (DUC and TAC), but are relatively small (in
    the range of a few hundred examples) for training neural models.
    In an attempt to drive research further, \citet{liu2018generating}
    tap into the potential of Wikipedia and propose a methodology for
    creating a large-scale dataset (WikiSum) for multi-document
    summarization with hundreds of thousands of instances. Wikipedia
    articles, specifically lead sections, are viewed as summaries of
    various topics indicated by their title, e.g.,``Florence'' or
    ``Natural Language Processing''. Documents cited in the Wikipedia
    articles or web pages returned by Google (using the section titles
    as queries) are seen as the source cluster which the lead section
    purports to summarize.
    
    
    Aside from the difficulties in obtaining training data, a major
    obstacle to the application of end-to-end models to multi-document
    summarization is the sheer size and number of source documents which
    can be very large. As a result, it is practically infeasible (given
    memory limitations of current hardware) to train a model which encodes
    all of them into vectors and subsequently generates a summary from
    them.  \citet{liu2018generating} propose a two-stage architecture, where
    an \emph{extractive} model first selects a subset of salient passages,
    and subsequently an \emph{abstractive} model generates the summary
    while conditioning on the extracted subset.  The selected passages are
    concatenated into a flat sequence and the Transformer
    \citep{vaswani2017attention}, an architecture well-suited to language
    modeling over long sequences, is used to decode the summary.
    
    Although the model of \citet{liu2018generating} takes an important  first step towards abstractive multi-document summarization, it 
    still considers the multiple input documents as a concatenated flat sequence, being agnostic 
    of the hierarchical structures and the relations that might exist among  documents.
    For
    example, different web pages might repeat the same content, include
    additional content, present contradictory information, or discuss the
    same fact in a different light \cite{radev:2000:SIGDIAL}. The
    realization that cross-document links are important in isolating
    salient information, eliminating redundancy, and creating overall
    coherent summaries, has led to the widespread adoption of graph-based
    models for multi-document summarization
    \cite{erkan2004lexrank,christensen2013towards,wan:2008:EMNLP2,parveen-strube:2014:TextGraphs-9}. 
    Graphs
    conveniently capture the relationships between textual units within a
    document collection and can be easily constructed under the assumption
    that text spans represent graph nodes and edges are semantic links
    between them.
    
    In this paper, we develop a neural summarization model which can
    effectively process multiple input documents and distill abstractive
    summaries. 
    Our model augments the previously proposed Transformer
    architecture with the ability to
    encode multiple documents in a hierarchical manner.
    We represent cross-document relationships via an
    attention mechanism which allows to share information across multiple
    documents as opposed to simply concatenating text spans and feeding
    them as a flat sequence to the model. 
    In this way, the model automatically \emph{learns} richer structural dependencies among
    textual units, thus incorporating well-established insights from
    earlier work. 
    Advantageously, the proposed
    architecture can easily benefit from information external to
    the model, i.e.,~by replacing inter-document attention with a
    graph-matrix computed based on the basis of lexical similarity
    \cite{erkan2004lexrank} or discourse relations
    \cite{christensen2013towards}.
    
    %
    %
    %
    %
    
    We evaluate our model on the WikiSum dataset and show experimentally that the proposed
    architecture brings substantial improvements over several strong
    baselines. We also find that the addition of a simple ranking
    module which scores documents based on their usefulness for the
    target summary can greatly boost the performance of a
    multi-document summarization system.
    
    
    \section{Related Work}
    \label{sec:related-work}
    
    Most previous multi-document summarization methods are extractive
    operating over graph-based representations of sentences or
    passages. Approaches vary depending on how edge weights are computed
    e.g., based on cosine similarity with \mbox{tf-idf} weights for words
    \cite{erkan2004lexrank} or on discourse relations
    \cite{christensen2013towards}, and the specific algorithm adopted for
    ranking text units for inclusion in the final summary. Several
    variants of the PageRank algorithm have been adopted in the literature
    \cite{erkan2004lexrank} in order to compute the
    importance or salience of a passage recursively based on the entire
    graph. More recently, \citet{yasunaga-EtAl:2017:CoNLL} propose a neural
    version of this framework, where salience is estimated using features
    extracted from sentence embeddings and graph convolutional networks
    \cite{kipf2016semi} applied over the relation graph representing
    cross-document links.

    Abstractive approaches have met with limited success. A few
    systems generate summaries based on sentence fusion, a technique
    which identifies fragments conveying common information across
    documents and combines these into sentences
    \cite{Barzilay:McKeown:2005,filippova-strube:2008:EMNLP,bing-EtAl:2015:ACL-IJCNLP}. Although
    neural abstractive models have achieved promising results on
    single-document summarization
    \cite{see-acl17,paulus2017deep,gehrmann2018bottom,asli-multiagent18}, the
    extension of sequence-to-sequence architectures to multi-document
    summarization is less straightforward. Apart from the lack of
    sufficient training data, neural models also face the
    computational challenge of processing multiple source
    documents. Previous solutions include model transfer
    \cite{W18-6545,D18-1387}, where a sequence-to-sequence model is
    pretrained on single-document summarization data and fine-tuned on
    DUC (multi-document) benchmarks, or unsupervised models relying on
    reconstruction objectives
    \cite{C16-1143,DBLP:journals/corr/abs-1810-05739}.
    
    \citet{liu2018generating} propose a methodology for constructing
    large-scale summarization datasets and a two-stage model which
    first extracts salient information from source documents and then
    uses a decoder-only architecture (that can attend to very long
    sequences) to generate the summary. We follow their setup in
    viewing multi-document summarization as a supervised machine
    learning problem and for this purpose assume access to large,
    labeled datasets (i.e.,~source documents-summary pairs).
    In
    contrast to their approach, we use a learning-based ranker and our abstractive model  can hierarchically encode the input documents, with the ability to 
    learn latent
    relations across documents and additionally incorporate 
    information encoded in well-known graph representations.

    \section{Model Description}
    \label{sec:problem-formulation}
    
    We follow \citet{liu2018generating} in treating the generation of lead
    Wikipedia sections as a multi-document summarization task. 
    The input to a hypothetical system is the title of a Wikipedia article
    and a collection of source documents, while the output is
    the Wikipedia article's first section.
    Source documents are webpages cited in the References section of
    the Wikipedia article and the top $10$~search results returned by
    Google (with the title of the article as the query).  Since source
    documents could be relatively long, they are split into multiple
    paragraphs by line-breaks. More formally, given title~$T$, and
    $L$~input paragraphs $\{P_1,\cdots,P_L \}$ (retrieved from
    Wikipedia citations and a search engine), the task is to generate
    the lead section~$D$ of the Wikipedia article.

    \begin{figure}[t]
        \centering
        \includegraphics[width=3in]{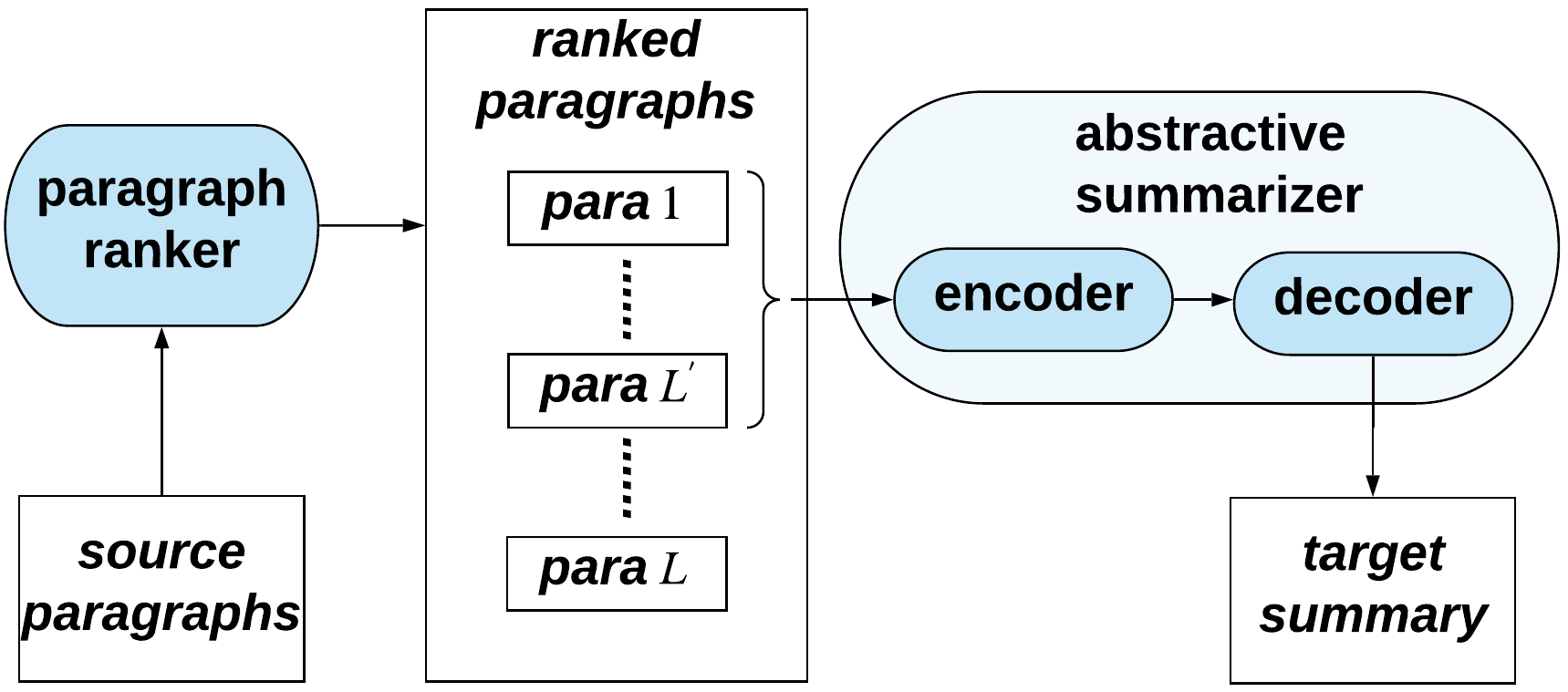}
        \caption{Pipeline of our multi-document summarization
            system. $L$~source paragraphs are first ranked and the
            $L'$-best ones serve as input to an encoder-decoder model
            which generates the  target summary.}
        \label{fig:overview}
    \end{figure}
    
    
    
    Our summarization system is illustrated in
    Figure~\ref{fig:overview}.  Since the input paragraphs are
    numerous and possibly lengthy, instead of directly applying an
    abstractive system, we first rank them and summarize the $L'$-best
    ones.  Our summarizer follows the very successful encoder-decoder
    architecture \cite{bahdanau2014neural}, where the encoder encodes
    the input text into hidden representations and the decoder
    generates target summaries based on these representations. In this
    paper, we focus exclusively on the encoder part of the model, our
    decoder follows the Transformer architecture introduced
    in~\citet{vaswani2017attention}; it generates a summary token by
    token while attending to the source input. We also use beam search
    and a length penalty~\citep{wu2016google} in the decoding process
    to generate more fluent and longer summaries.
    
    \subsection{Paragraph Ranking}
    \label{sec:paragraph-preranking}
    
    Unlike \citet{liu2018generating} who rank paragraphs based on
    their similarity with the title (using \mbox{tf-idf}-based cosine
    similarity), we adopt a learning-based approach.  A logistic
    regression model is applied to each paragraph to calculate a score
    indicating whether it should be selected for summarization.  We
    use two recurrent neural networks with Long-Short Term Memory
    units (LSTM; \citealt{hochreiter:1997:nc}) to represent title~$T$
    and source paragraph~$P$:
    \begin{gather}
    \{u_{t1}, \cdots, u_{tm}\}=\mathrm{lstm}_t(\{w_{t1}, \cdots, w_{tm}\})\\
    \{u_{p1}, \cdots, u_{pn}\}=\mathrm{lstm}_p(\{w_{p1}, \cdots, w_{pn}\})
    \end{gather}
    where~$w_{ti}, w_{pj}$~are word embeddings for tokens in $T$ and $P$, and~$u_{ti},u_{pj}$ are the updated vectors for each token after applying the
    LSTMs.   
    
    A max-pooling operation is then used over title vectors to obtain
    a fixed-length representation~$\hat{u}_t$:
    \begin{gather}
    \hat{u}_t=\mathrm{maxpool}(\{u_{t1}, \cdots, u_{tm}\})
    \end{gather}
    We concatenate~$\hat{u}_t$ with the vector $u_{pi}$ of each token in the paragraph
    and apply a non-linear transformation to extract features for matching
    the title and the paragraph. A second max-pooling operation yields
    the final paragraph vector~$\hat{p}$:
    \begin{gather}
    p_i = \tanh(W_1([u_{pi};\hat{u}_t])) \\
    \hat{p} =\mathrm{maxpool}(\{p_{1}, \cdots, p_{n}\})
    \end{gather}
    Finally, to estimate whether a paragraph should be selected, we
    use a linear transformation and a sigmoid function:
    \begin{gather}
    s=\sigmoid(W_2\hat{(p)})
    \end{gather}
    where $s$ is the score indicating whether paragraph~$P$ should be
    used for summarization.
    
    All input paragraphs $\{P_1,\cdots,P_L \}$ receive scores
    $\{s_1,\cdots,s_L \}$.  The model is trained by minimizing the
    cross entropy loss between~$s_i$ and ground-truth scores~$y_i$
    denoting the relatedness of a paragraph to the gold standard
    summary. 
    We adopt ROUGE-2 recall (of paragraph~$P_i$ against gold
    target text~$D$) as $y_i$.  In testing, input paragraphs are
    ranked based on the model predicted scores and an ordering
    $\{R_1,\cdots,R_L\}$ is generated. The first $L'$~paragraphs
    $\{R_1,\cdots,R_{L'}\}$ are selected as input to the second
    abstractive stage.

    \subsection{Paragraph Encoding}
    \label{sec:abstr-summ-with}
    
    Instead of treating the selected paragraphs as a very long
    sequence, we develop a hierarchical model based on the Transformer architecture \cite{vaswani2017attention} to capture inter-paragraph relations. The model is composed of several
    \emph{local} and \emph{global} transformer layers which can be
    stacked freely.  Let~$t_{ij}$ denote the $j$-th token in the
    $i$-th ranked paragraph~$R_i$; 
    the model takes vectors $x^0_{ij}$
    (for all tokens) as input.  For the $l$-th transformer layer, the
    input will be $x^{l-1}_{ij}$, and the output is written as
    $x^l_{ij}$.

    \subsubsection{Embeddings}

    Input tokens are first represented by word embeddings.  Let~$w_{ij}\in \mathbb{R}^d$ denote the embedding assigned to~$t_{ij}$.
    Since the Transformer is a non-recurrent model, we also assign a
    special positional embedding $pe_{ij}$ to $t_{ij}$, to indicate
    the position of the token within the input.

    To calculate positional embeddings, we follow
    \citet{vaswani2017attention} and use sine and cosine functions of
    different frequencies. The embedding $e_p$ for
    the $p$-th element in a sequence is:
    \begin{gather}
    e_p[i] = \sin(p/10000^{2i/d})\\
    e_p[2i+1] = \cos(p/10000^{2i/d})
    \end{gather}
    where $e_p[i]$ indicates the $i$-th dimension of the embedding
    vector.  Because each dimension of the positional encoding
    corresponds to a sinusoid, for any fixed offset $o$, $e_{p+o}$ can
    be represented as a linear function of $e_{p}$, which enables the
    model to distinguish relative positions of input elements.
    
    In multi-document summarization,  token~$t_{ij}$ has two
    positions that need to be considered, namely~$i$ (the rank of the
    paragraph) and~$j$ (the position of the token within the
    paragraph).  Positional embedding~$pe_{ij} \in \mathbb{R}^d$ represents both
    positions (via concatenation) and is added to word
    embedding~$w_{ij}$ to obtain the final input vector $x^0_{ij}$ :
    \begin{gather}
    pe_{ij}=[e_i;e_j]\\
    x^0_{ij} = w_{ij}+ pe_{ij}
    \end{gather}

    \subsubsection{Local Transformer Layer}
    
    A local transformer layer is used to encode contextual information
    for tokens within each paragraph. The local transformer layer is
    the same as the vanilla transformer
    layer~\cite{vaswani2017attention}, and composed of two sub-layers:
    \begin{gather}
    h=\mathrm{LayerNorm}(x^{l-1}+\mathrm{MHAtt}(x^{l-1}))\\
    x^l=\mathrm{LayerNorm}(h+\mathrm{FFN}(h))
    \end{gather}
    where 
    $\mathrm{LayerNorm}$ is  layer normalization  proposed
    in~\citet{ba2016layer}; $\mathrm{MHAtt}$ is the multi-head
    attention mechanism introduced in~\citet{vaswani2017attention} which
    allows each token to attend to other tokens with different attention distributions; and $\mathrm{FFN}$
    is a two-layer feed-forward network with ReLU as hidden activation
    function.

    \subsubsection{Global Transformer Layer}
\begin{figure}
        \centering
        \includegraphics[width=3in]{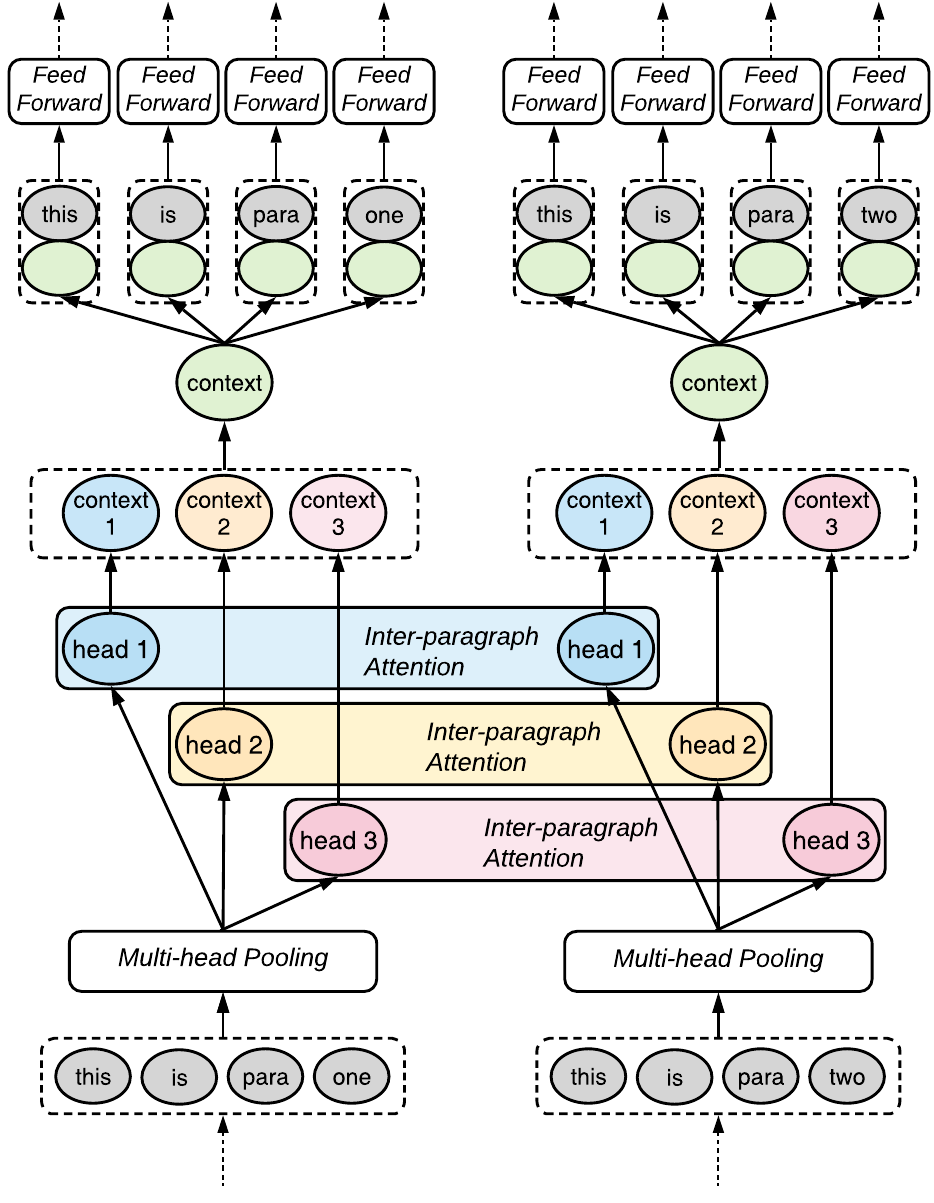}
        \caption{A global transformer layer. Different colors indicate
            different heads in multi-head pooling and inter-paragraph
            attention.}
        \label{fig:discourse}
    \end{figure}

    A global transformer layer is used to exchange information across
    multiple paragraphs.  As shown in Figure~\ref{fig:discourse}, we
    first apply a multi-head pooling operation to each
    paragraph. Different heads will encode paragraphs with different
    attention weights.  Then, for each head, an inter-paragraph
    attention mechanism is applied, where each paragraph can collect
    information from other paragraphs by self-attention, generating a
    context vector to capture contextual information from the whole
    input.  Finally, context vectors are concatenated, linearly
    transformed, added to the vector of each token, and fed to a
    feed-forward layer, updating the representation of each token with
    global information.

    \paragraph{Multi-head Pooling}
    To obtain fixed-length paragraph representations, we apply a
    weighted-pooling operation; instead of using only one representation
    for each paragraph, we introduce a multi-head pooling mechanism, where
    for each paragraph,  weight distributions over tokens are
    calculated, allowing the model to flexibly encode paragraphs in
    different representation subspaces by attending to different words.

    Let  $x^{l-1}_{ij}\in \mathbb{R}^{d}$ denote the output vector
    of the last transformer layer for token~$t_{ij}$, which is used
    as input for the current layer. 
    For each paragraph $R_i$, for head $z \in \{1,\cdots,n_{head}\}$, we
    first transform the input vectors into attention scores $a^z_{ij}
    $ and value vectors $b^z_{ij}$.  Then, for each head,
    we calculate a probability distribution $\hat{a}^z_{ij} $ over
    tokens within the paragraph based on attention scores:
    \begin{gather}
    a^z_{ij} = W^z_ax^{l-1}_{ij}\\
    b^z_{ij} = W^z_bx^{l-1}_{ij}\\
    \hat{a}^z_{ij} = exp(a^z_{ij} )/{\sum_{j=1}^nexp({a^z_{ij} }})
    \end{gather}
    where $W^z_a\in \mathbb{R}^{1*d}$ and $W^z_b\in
    \mathbb{R}^{d_{head}*d}$ are weights.  $d_{head}=d/n_{head}$ is
    the dimension of each head. $n$ is the number of tokens in $R_i$.
    
    We next apply a weighted summation with another linear
    transformation and layer normalization to obtain vector $head^z_i$
    for the paragraph:
    \begin{gather}
    head^z_i = \mathrm{LayerNorm}(W^z_c\sum_{j=1}^na^z_{ij}b^z_{ij})
    \end{gather}
    where $W^z_c \in \mathbb{R}^{d_{head}*d_{head}}$ is the weight.
    
    The model can flexibly incorporate multiple heads, with each
    paragraph having multiple attention distributions, thereby
    focusing on different views of the input.

    \paragraph{Inter-paragraph Attention}
    We model the dependencies across multiple paragraphs with an
    inter-paragraph attention mechanism. Similar to self-attention, inter-paragraph attention allows for each paragraph to
    attend to other paragraphs by calculating an attention distribution:
    \begin{gather}
    q^z_i = W^z_qhead^z_i\\
    k^z_i = W^z_khead^z_i\\
    v^z_i = W^z_vhead^z_i\\
    context^z_i =\sum_{i=1}^m \frac{exp({q^z_i}^Tk^z_{i'})}{\sum_{o=1}^m{exp({q^z_i}^Tk^z_{o})}}v^z_{i'} \label{eq:context}
    \end{gather}
    where $q^z_i,k^z_i, v^z_i \in \mathbb{R}^{d_{head}*d_{head}}$ are
    query, key, and value vectors that are linearly transformed from
    $head^z_i$ as in~\citet{vaswani2017attention};
    $context^z_i\in \mathbb{R}^{d_{head}}$ represents the context
    vector generated by a self-attention operation over all
    paragraphs. $m$ is the number of input paragraphs.
    Figure~\ref{fig:discourse} provides a schematic view of
    inter-paragraph attention.
    
    \paragraph{Feed-forward Networks}
    We next update token representations with contextual information.
    We first fuse information from all heads by concatenating all
    context vectors and applying a linear transformation with weight
    $W_c \in \mathbb{R}^{d*d}$:
    \begin{gather}
    c_i=W_c[context^1_i; \cdots; context^{n_{head}}_i]
    \end{gather}
    
    We then add~$c_i$ to each input token vector~$x^{l-1}_{ij}$, and feed it
    to a two-layer feed-forward network with ReLU as the activation
    function and a highway layer normalization on top:
    \begin{gather}
    g_{ij} = W_{o2} \mathrm{ReLU}(W_{o1}(x^{l-1}_{ij}+c_i))\\
    x^l_{ij} =  \mathrm{LayerNorm}(g_{ij }+x^{l-1}_{ij})
    \end{gather}
    where $W_{o1} \in \mathbb{R}^{d_{ff}*d}$ and $W_{o2} \in
    \mathbb{R}^{d*d_{ff}}$ are the weights, $d_{ff}$ is the hidden  size of the feed-forward later. 
    This way, each token
    within paragraph $R_i$ can collect information from other
    paragraphs in a hierarchical and efficient manner.

    \subsubsection{Graph-informed Attention}
    \label{sec:graph-inform-attent}
    
    The inter-paragraph attention mechanism can be viewed as learning
    a latent graph representation (self-attention weights) of the
    input paragraphs. Although previous work has shown that similar
    latent representations are beneficial for down-stream NLP tasks
    \cite{TACL1185,kim2017structured,Williams:ea:2018,Niculae:ea:2018,fernandes2018structured},
    much work in multi-document summarization has taken advantage of
    explicit graph representations, each focusing on different facets
    of the summarization task (e.g.,~capturing redundant information
    or representing passages referring to the same event or entity).
    One advantage of the hierarchical transformer is that we can
    easily incorporate graphs external to the model, to generate
    better summaries.
    
    We experimented with two well-established graph representations
    which we discuss briefly below. However, there is nothing inherent
    in our model that restricts us to these, any graph modeling
    relationships across paragraphs could have been used instead. Our
    first graph aims to capture lexical relations; graph nodes
    correspond to paragraphs and edge weights are cosine similarities
    based on tf-idf representations of the paragraphs.  Our second
    graph aims to capture discourse relations
    \cite{christensen2013towards}; it builds an Approximate Discourse
    Graph (ADG)~\citep{yasunaga-EtAl:2017:CoNLL} over paragraphs;
    edges between paragraphs are drawn by counting (a)~co-occurring
    entities and (b)~discourse markers (e.g., \textsl{however},
    \textsl{nevertheless}) connecting two adjacent paragraphs (see the
    Appendix for details on how ADGs are constructed).
    

    We represent such graphs with a matrix~$G$, where~$G_{ii'}$ is the
    weight of the edge connecting paragraphs~$i$ and~$i'$. We can then
    inject this graph into our hierarchical transformer by simply
    substituting one of its (learned) heads $z'$
    with~$G$. Equation~\eqref{eq:context} for calculating the context
    vector for this head is modified as:
    \begin{gather}
    context^{z'}_i = \sum_{i'=1}^m\frac{G_{ii'}}{\sum_{o=1}^mG_{io}}v^{z'}_{i'}
    \end{gather}

    \section{Experimental Setup}
    \label{sec:experiments}
    
    
    
    \paragraph{WikiSum Dataset}
    We used the scripts and urls provided in~\citet{liu2018generating}
    to crawl Wikipedia articles and source reference documents. We
    successfully crawled $78.9$\% of the original documents (some urls
    have become invalid and corresponding documents could not be
    retrieved).  We further removed clone paragraphs (which are exact
    copies of some parts of the Wikipedia articles); these were
    paragraphs in the source documents whose bigram recall against the
    target summary was higher than~$0.8$.  On average, each input
    has~$525$ paragraphs, and each paragraph has~$70.1$ tokens.  The
    average length of the target summary is~$139.4$ tokens.  We split
    the dataset with $1,579,360$ instances for training, $38,144$ for
    validation and $38,205$ for test.
    
    For both ranking and summarization stages, we encode source
    paragraphs and target summaries using subword tokenization with
    SentencePiece~\citep{kudo2018sentencepiece}. Our vocabulary
    consists of~$32,000$ subwords and is shared for both source and
    target.

    \paragraph{Paragraph Ranking}

    To train the regression model, we calculated the ROUGE-2
    recall~\citep{lin:2004:ACLsummarization} of each paragraph against
    the target summary and used this as the ground-truth score.  The
    hidden size of the two LSTMs was set to~$256$, and dropout (with
    dropout probability of~$0.2$) was used before all linear layers.
    Adagrad~\cite{duchi2011adaptive} with learning rate $0.15$ is used
    for optimization.  We compare our ranking model against the method
    proposed in \citet{liu2018generating} who use the tf-idf cosine
    similarity between each paragraph and the article title to rank
    the input paragraphs.  We take the first $L'$~paragraphs from the
    ordered paragraph set produced by our ranker and the
    similarity-based method, respectively. We concatenate these
    paragraphs and calculate their ROUGE-L recall against the gold
    target text. The results are shown in Table~\ref{pre}. We can see
    that our ranker effectively extracts related paragraphs and
    produces more informative input for the downstream summarization
    task.

    \begin{table}[t]
        \centerline{
            \begin{tabular}{@{}l@{~}|@{~}l@{~~~}l@{~~~}l@{~~}l@{}}
                \thickhline
                \multirow{2}{*}{Methods} & \multicolumn{4}{c}{ROUGE-L Recall} \\
                & $\displaystyle{L'=5}$     & $\displaystyle{L'=10}$   & $\displaystyle{L'=20}$   & $\displaystyle{L'=40}$   \\ \thickhline
                Similarity      & 24.86   &32.43  & 40.87 & 49.49  \\
                Ranking      & 39.38  & 46.74  & 53.84 & 60.42 \\\thickhline
            \end{tabular}
        }
        \caption{ROUGE-L recall against target summary for \mbox{$L'$-best}
            paragraphs obtained with tf-idf cosine similarity and our ranking
            model. } 
        \label{pre}
    \end{table}

    \begin{table*}[t]
        \center
        \begin{tabular}{lccc} \thickhline
            Model & ROUGE-1 & ROUGE-2 & ROUGE-L \\\thickhline
            Lead&38.22&16.85&26.89\\ 
            LexRank&36.12&11.67&22.52\\ \hline
            FT (600 tokens, no ranking)&35.46&20.26&30.65\\
            FT (600 tokens)&40.46&25.26&34.65\\
            FT (800 tokens)&40.56&25.35&34.73\\
            FT (1,200 tokens)&39.55&24.63&33.99\\
            T-DMCA (3000 tokens)&40.77&25.60&34.90\\\hline
            HT (1,600 tokens)&\textbf{40.82}&\textbf{25.99}&35.08\\
            HT (1,600 tokens) + Similarity Graph&40.80&25.95&35.08\\
            HT (1,600 tokens) + Discourse Graph&40.81&25.95&\textbf{35.24}\\
            \hline
            HT (train on  1,600 tokens/test on 3000 tokens)  &\textbf{41.53}&\textbf{26.52}&\textbf{35.76} \\\thickhline

        \end{tabular}        
        \caption{\label{tab:automatic} Test set results on  the WikiSum
            dataset using  ROUGE $F_1$. 
        }
    \end{table*}
    
    
    \paragraph{Training Configuration}
    
    In all abstractive models, we apply dropout (with probability
    of~$0.1$) before all linear layers; label
    smoothing~\citep{szegedy2016rethinking} with smoothing
    factor~$0.1$ is also used.  Training is in traditional
    sequence-to-sequence manner with maximum likelihood estimation.
    The optimizer was Adam~\cite{kingma2014adam} with learning rate of
    $2, \beta_1 = 0.9$, and $\beta_2 = 0.998$; we also applied
    learning rate warmup over the first $8,000$ steps, and decay as in~\cite{vaswani2017attention}.  All transformer-based models had $256$~hidden units; the
    feed-forward hidden size was~$1,024$ for all layers.  All models
    were trained on $4$ GPUs (NVIDIA TITAN Xp) for $500,000$ steps. We
    used gradient accumulation to keep training time for all models
    approximately consistent.  We selected the $5$ best checkpoints
    based on performance on the validation set and report averaged
    results on the test set.
    
    During decoding we use beam search with beam size~$5$ and length
    penalty with~$\alpha = 0.4$ \citep{wu2016google}; we decode until
    an end-of-sequence token is reached.

    \paragraph{Comparison Systems}
    We compared the proposed hierarchical transformer against several
    strong baselines:
    \begin{itemize}
        
        \item[\textbf{Lead}] is a simple baseline that concatenates the title
        and ranked paragraphs, and extracts the first~$k$ tokens; we set~$k$ to
        the length of the ground-truth target.

        \item[\textbf{LexRank}] \cite{erkan2004lexrank} is a widely-used
        graph-based extractive summarizer; we build a graph with
        paragraphs as nodes and edges weighted by tf-idf cosine
        similarity; we run a PageRank-like algorithm on this graph to
        rank   and select paragraphs
        until the length of the ground-truth summary
        is reached.

        \item[\textbf{Flat Transformer (FT)}] is a baseline that applies
        a Transformer-based encoder-decoder model to a flat token
        sequence. We used a $6$-layer transformer. The title and ranked
        paragraphs were concatenated and truncated to~$600, 800$, and
        $1,200$~tokens.

        \item[\textbf{T-DMCA}] is the best performing model
        of~\citet{liu2018generating} and a shorthand for Transformer
        Decoder with Memory Compressed Attention; they only used a
        Transformer decoder and compressed the key and value in
        self-attention with a convolutional layer. The model has $5$~layers as in~\citet{liu2018generating}. Its hidden size is $512$ and its
        feed-forward hidden size is~$2,048$.  The title and ranked
        paragraphs were concatenated and truncated to~3,000~tokens.

        \item[\textbf{Hierarchical Transformer (HT)} ] is the model
        proposed in this paper.  The model
        architecture is a $7$-layer network (with $5$ local-attention
        layers at the bottom and $2$ global attention layers at the
        top).  The model takes the title and $L'=24$ paragraphs as input to produce a
        target summary, which leads to approximately $1,600$ input tokens
        per instance.
        
    \end{itemize}

    \section{Results}
    \label{sec:results}
    
    \paragraph{Automatic Evaluation}
    
    We evaluated summarization quality using ROUGE
    $F_1$~\citep{lin:2004:ACLsummarization}. We report unigram and
    bigram overlap (\mbox{ROUGE-1} and \mbox{ROUGE-2}) as a means of
    assessing informativeness and the longest common subsequence
    (ROUGE-L) as a means of assessing fluency.
    
    Table~\ref{tab:automatic} summarizes our results. The first block
    in the table includes extractive systems (Lead, LexRank), the
    second block includes several variants of Flat Transformer-based
    models (FT, \mbox{T-DMCA}), while the rest of the table presents
    the results of our Hierarchical Transformer (HT). As can be seen,
    abstractive models generally outperform extractive ones.  The Flat
    Transformer, achieves best results when the input length is set
    to~$800$ tokens, while longer input (i.e., $1,200$ tokens) actually
    hurts performance.  The Hierarchical Transformer with $1,600$ input
    tokens, outperforms FT, and even T-DMCA when the latter is
    presented with~$3,000$ tokens.  Adding an external graph also seems
    to help the summarization process. The similarity graph does not
    have an obvious influence on the results, while the discourse
    graph boosts ROUGE-L by~$0.16$.
    
    We also found that the performance of the Hierarchical Transformer further
    improves when the model is presented with longer input at test
    time.\footnote{This was not the case with the other Transformer
        models.}  As shown in the last row of Table~\ref{tab:automatic},
    when testing on $3,000$ input tokens, summarization quality improves
    across the board.  This suggests that the model can potentially
    generate better summaries without increasing training time.
    

    \begin{table}[]
        \center
        \begin{tabular}{lccc} \thickhline
            Model                                 & R1                    & R2 & RL \\ \thickhline
            HT                                    &      40.82                 & 25.99   & 35.08   \\ 
            HT w/o PP            & 40.21& 24.54   & 34.71   \\
            
            HT w/o MP & 39.90& 24.34   & 34.61    \\
            HT w/o GT     &    39.01                   & 22.97  & 33.76   \\ \thickhline
        \end{tabular} 
        \caption{\label{tab:ablation} Hierarchical Transformer and
            versions thereof without (w/o)  paragraph position (PP),
            multi-head pooling (MP), and  global transformer layer (GT).}
    \end{table}

    Table~\ref{tab:ablation} summarizes  ablation studies
    aiming to assess the contribution of individual 
    components. Our experiments confirmed that encoding paragraph
    position in addition to token position within each paragraph is
    beneficial (see row w/o PP), as well as multi-head pooling (w/o MP
    is a model where the number of heads is set to~$1$), and the global
    transformer layer (w/o GT is a model with only $5$~local transformer
    layers in the encoder).
    
    \paragraph{Human Evaluation}
    
    In addition to automatic evaluation, we also assessed system
    performance by eliciting human judgments on $20$ randomly selected
    test instances.  Our first evaluation study quantified the degree
    to which summarization models retain key information from the
    documents following a question-answering (QA) paradigm
    ~\citep{clarke2010discourse,narayan2018ranking}.  We created a set
    of questions based on the gold summary under the assumption that
    it contains the most important information from the input
    paragraphs.  We then examined whether participants were able to
    answer these questions by reading system summaries alone without
    access to the gold summary. The more questions a system can
    answer, the better it is at summarization.  We created $57$
    questions in total varying from two to four questions per gold
    summary.  Examples of questions and their answers are given in
    Table~\ref{tab:output}.  We adopted the same scoring mechanism
    used in~\citet{clarke2010discourse}, i.e., correct answers are
    marked with~$1$, partially correct ones with~$0.5$, and~$0$ otherwise.
    A system's score is the average of all question scores.

    Our second evaluation study assessed the overall quality of the
    summaries by asking participants to rank them taking into account
    the following criteria: \emph{Informativeness} (does the summary
    convey important facts about the topic in question?),
    \emph{Fluency} (is the summary fluent and grammatical?), and
    \emph{Succinctness} (does the summary avoid repetition?).  We used
    Best-Worst Scaling \cite{louviere2015best}, a less labor-intensive
    alternative to paired comparisons that has been shown to produce
    more reliable results than rating scales~\cite{bestworstscaling}.
    Participants were presented with the gold summary and summaries
    generated from $3$ out of $4$ systems and were asked to decide which
    summary was the best and which one was the worst in relation to the
    gold standard, taking into account the criteria mentioned above.
    The rating of each system was computed as the percentage of times
    it was chosen as best minus the times it was selected as
    worst. Ratings range from $-1$ (worst) to $1$
    (best).  

    \begin{table}[t]
        
        \center
        \begin{tabular}{lcr} \thickhline
            Model  &  \multicolumn{1}{c}{QA} & \multicolumn{1}{c}{Rating}  \\\thickhline
            Lead      & 31.59     &    -0.383    \\
            FT        &   35.69   &    0.000     \\
            T-DMCA    &   43.14   &     0.147   \\
            HT        &\bf{54.11} &   \bf{0.237}  \\ \thickhline
        \end{tabular}
        \caption{\label{tab:humans} System scores based on questions
            answered by AMT participants and  summary quality rating.} 
    \end{table}
    
    Both evaluations were conducted on the Amazon Mechanical Turk
    platform with $5$ responses per hit.  Participants evaluated
    summaries produced by the Lead baseline, the Flat Transformer,
    T-DMCA, and our {Hierarchical Transformer}.  All evaluated systems
    were variants that achieved the best performance in automatic
    evaluations.  As shown in Table~\ref{tab:humans}, on both
    evaluations, participants overwhelmingly prefer our model (HT).
    All pairwise comparisons among systems are statistically
    significant (using a one-way ANOVA with post-hoc Tukey HSD tests;
    \mbox{$p < 0.01$}). Examples of system output are provided in
    Table~\ref{tab:output}.

    \begin{table*}[t!]
        \small
        \begin{center}
            \begin{tabular}{@{~}lp{14cm}@{}}
                \multicolumn{2}{c}{\normalsize Pentagoet Archeological District} \\            
                
                \thickhline            \begin{sideways}\hspace*{-1ex}\textsc{Gold}\end{sideways}
                & \multicolumn{1}{l}{\begin{tabularx}{14cm}[c]{@{}X@{}}The Pentagoet Archeological District is a National Historic Landmark District located at the southern edge of the Bagaduce Peninsula in Castine, Maine.   It is the site of Fort Pentagoet, a 17th-century fortified trading post established by fur traders of French Acadia.   From 1635 to 1654 this site was a center of trade with the local Abenaki, and marked the effective western border of Acadia with New England.   From 1654 to 1670 the site was under English control, after which it was returned to France by the Treaty of Breda.   The fort was destroyed in 1674 by Dutch raiders.   The site was designated a National Historic Landmark in 1993.   It is now a public park.\end{tabularx}}
                \\ 
                \hline            \begin{sideways}\hspace*{-1ex}QA\end{sideways}
                & \begin{tabularx}{14cm}[c]{@{}X@{}}
                    
                    What is the Pentagoet Archeological District?\hfill [a National Historic Landmark District]\\
                    Where is it located?\hfill [Castine , Maine]\\
                    What did the Abenaki Indians use the site for?\hfill [trading center]
                \end{tabularx}
                \\
                \hline           \begin{sideways}\hspace*{-2ex}\textsc{Lead}\end{sideways}
                & \begin{tabularx}{14cm}[c]{@{}X@{}}The Pentagoet Archeological District is a National Historic Landmark District located in Castine, Maine. This district forms part of the traditional homeland of the Abenaki Indians, in particular the Penobscot tribe. In the colonial period, Abenakis frequented the fortified trading post at this site, bartering moosehides, sealskins, beaver and other furs in exchange for European commodities. 
                    "Pentagoet Archeological district" is a National Historic Landmark District located at the southern edge of the Bagaduce Peninsula in Treaty Of Breda.
                \end{tabularx}
                \\ 
                \hline      \begin{sideways}\hspace*{-1ex}\textsc{FT}\end{sideways}
                & \begin{tabularx}{11cm}[c]{@{}X@{}}the Pentagoet Archeological district is a National Historic Landmark District located at the southern edge of the Bagaduce Peninsula in Treaty Of Breda.  It was listed on the national register of historic places in 1983.\end{tabularx}
                \\
                \hline      \begin{sideways}\hspace*{-4ex}\textsc{T-DMCA}\end{sideways}
                & \begin{tabularx}{14cm}[c]{@{}X@{}}The Pentagoet Archeological District is a national historic landmark district located in castine , maine .  this district forms part of the traditional homeland of the abenaki indians , in particular the Penobscot tribe.  The district was listed on the national register of historic places in 1982.\\
                \end{tabularx}

                \\
                \hline  \begin{sideways}\hspace*{-1ex}\textsc{HT}\end{sideways}
                & \begin{tabularx}{14cm}[c]{@{}X@{}}The Pentagoet Archeological district is a National Historic Landmark District located in Castine, Maine.  This district forms part of the traditional homeland of the Abenaki Indians, in particular the Penobscot tribe.  In the colonial period, Abenaki frequented the fortified trading post at this site, bartering moosehides, sealskins, beaver and other furs in exchange for European commodities.\end{tabularx}
                \\ \thickhline                                                    \multicolumn{2}{c}{}\\
                \multicolumn{2}{c}{\normalsize Melanesian Whistler}\\
                \thickhline            \begin{sideways}\hspace*{-2ex}\textsc{Gold}\end{sideways}
                & \multicolumn{1}{l}{\begin{tabularx}{14cm}[c]{@{}X@{}}The Melanesian whistler or Vanuatu whistler (Pachycephala chlorura) is a species of passerine bird in the whistler family Pachycephalidae.   It is found on the Loyalty Islands, Vanuatu, and Vanikoro in the far south-eastern Solomons.\end{tabularx}}
                \\ 
                \hline            \begin{sideways}\hspace*{-1ex}QA\end{sideways}
                & \begin{tabularx}{14cm}[c]{@{}X@{}}
                    What is the Melanesian Whistler? \hfill[a species of passerine bird in the whistler family Pachycephalidae]\\
                    Where is it found? \hfill[Loyalty Islands , Vanuatu , and Vanikoro in the far south-eastern Solomons]	
                \end{tabularx}
                \\
                \hline           \begin{sideways}\hspace*{-2ex}\textsc{Lead}\end{sideways}
                & \begin{tabularx}{14cm}[c]{@{}X@{}}The Australian golden whistler (Pachycephala pectoralis) is a species of bird found in forest, woodland, mallee, mangrove and scrub in Australia (except the interior and most of the north) Most populations are resident, but some in south-eastern Australia migrate north during the winter.
                \end{tabularx}
                \\ 
                \hline      \begin{sideways}\hspace*{-1ex}\textsc{FT}\end{sideways}
                & \begin{tabularx}{14cm}[c]{@{}X@{}}The Melanesian whistler (P. Caledonica) is a species of bird in the family Muscicapidae.  It is endemic to Melanesia.\end{tabularx}
                \\
                \hline      \begin{sideways}\hspace*{-4ex}\textsc{~~T-DMCA}\end{sideways}
                & \begin{tabularx}{14cm}[c]{@{}X@{}}The Australian golden whistler (Pachycephala chlorura) is a species of bird in the family Pachycephalidae, which is endemic to Fiji.\\\\\end{tabularx}
                \\
                \hline  \begin{sideways}\hspace*{-1ex}\textsc{HT}\end{sideways}
                & \begin{tabularx}{14cm}[c]{@{}X@{}}The Melanesian whistler (Pachycephala chlorura) is a species of bird in the family Pachycephalidae, which is endemic to Fiji.\end{tabularx}
                \\ \thickhline                                                                     
            \end{tabular}
        \end{center}
        \caption{\label{tab:output} \textsc{Gold} human authored summaries, questions based on
            them (answers shown in square brackets) and automatic summaries  produced by the \textsc{Lead-3} baseline,
            the Flat Transformer~(\textsc{FT}), 
            \textsc{T-DMCA}~\cite{ liu2018generating} and our
            Hierachical Transformer (\textsc{HT}).\label{tab:system-output}}
    \end{table*}

    \section{Conclusions}
    
    In this paper we conceptualized abstractive multi-document
    summarization as a machine learning problem. We proposed a new
    model which is able to encode multiple input documents hierarchically,
    learn latent relations across them, and
    additionally incorporate structural information from well-known
    graph representations. We have also demonstrated the importance of
    a learning-based approach for selecting which documents to
    summarize. Experimental results show that our model produces
    summaries which are both fluent and informative outperforming
    competitive systems by a wide margin. In the future we would like
    to apply our hierarchical transformer to question answering and
    related textual inference tasks.

    \section*{Acknowledgments}
    We would like to thank Laura Perez-Beltrachini for her help with
    preprocessing the  dataset.  
    This research is
    supported by a Google PhD Fellowship to the first author.
    The authors gratefully acknowledge
    the financial support of the European Research Council (award number
    681760).

    \newpage
    \bibliographystyle{acl_natbib}
    \bibliography{acl2019}
    
    \appendix
    
    \section{Appendix} 
    
    We describe here how the similarity and discourse graphs discussed in
    Section~\ref{sec:graph-inform-attent} were created. These graphs were
    added to the hierarchical transformer model as a means to enhance
    summary quality (see Section~\ref{sec:results} for details).
    \subsection{Similarity Graph}
    The similarity graph $S$ is based on tf-idf cosine similarity. The
    nodes of the graph are paragraphs. We first represent each
    paragraph~$p_i$ as a bag of words. Then, we calculate the tf-idf
    value~$v_{ik}$ for each token $t_{ik}$ in a paragraph:
    
    \begin{gather}
    v_{ik} = N_w(t_{ik})log(\frac{N_d}{N_{dw}(t_{ik})})
    \end{gather}
    where $Nw(t)$ is the count of word~$t$ in the paragraph, $N_d$ is
    the total number of paragraphs, and~$N_{dw}(t)$ is the total
    number of paragraphs containing the word. We thus obtain a tf-idf
    vector for each paragraph. Then, for all paragraph pairs
    \mbox{$<p_i, p_{i'}>$}, we calculate the cosine similarity of
    their tf-idf vectors and use this as the weight $S_{i{i'}}$ for
    the edge connecting the pair in the graph.  We remove edges with
    weights lower than~$0.2$.
    
    \subsection{Discourse Graphs}
    To build the Approximate Discourse Graph (ADG)~$D$, we
    follow~\citet{christensen2013towards} and
    \citet{yasunaga-EtAl:2017:CoNLL}. The original ADG makes use of
    several complex features. Here, we create a simplified version
    with only two features (nodes in this graph are again paragraphs).

    \paragraph{Co-occurring Entities} For each paragraph $p_i$, we
    extract a set of entities $E_i$ in the paragraph using the
    Spacy\footnote{https://spacy.io/api/entityrecognizer} NER
    recognizer. We only use entities with type \{\texttt{PERSON, NORP,
        FAC, ORG, GPE, LOC, EVENT, WORK\_OF\_ART, LAW}\}. For each
    paragraph pair \mbox{$<p_i, p_j>$}, we count $e_{ij}$, the number
    of entities with exact match.

    \paragraph{Discourse Markers}
    We use the following $36$ explicit discourse markers to identify edges between
    two adjacent paragraphs in a source webpage:
    \begin{quote}
        again, also, another, comparatively, furthermore, at the
        same time,however, immediately, indeed, instead, to be sure,
        likewise, meanwhile, moreover, nevertheless, nonetheless, notably,
        otherwise, regardless, similarly, unlike, in addition, even, in
        turn, in exchange, in this case, in any event, finally, later, as
        well, especially, as a result, example, in fact, then, the day
        before
    \end{quote} 
    If two paragraphs \mbox{$<p_i, p_{i'}>$} are adjacent in one
    source webpage and they are connected with one of the above~$36$
    discourse markers, $m_{ii'}$ will be $1$, otherwise it will be $0$.
    

    The final edge weight $D_{ii'}$ is the weighted sum of $e_{ii'}$ and
    $m_{ii'}$
    \begin{gather}
    D_{ii'} = 0.2*e_{ii'}+m_{ii'}
    \end{gather}

\end{document}